\documentclass[letterpaper, 10 pt, journal, twoside]{IEEEtran}  

\IEEEoverridecommandlockouts                              

\usepackage{graphicx} 
\usepackage{amsmath} 
\usepackage{amssymb}  
\usepackage{subcaption}
\usepackage{comment}
\usepackage[font=footnotesize]{caption}
\usepackage[capitalise]{cleveref}
\usepackage[abbreviations]{glossaries-extra}
\usepackage{siunitx}
\usepackage{booktabs}

\title{An Efficient Multi-Robot Arm Coordination Strategy for Pick-and-Place Tasks using Reinforcement Learning}

\author{Tizian Jermann, Hendrik Kolvenbach, Fidel Esquivel Estay, Koen Kr{\"a}mer and Marco Hutter
\thanks{All authors are with ETH Zurich, Robotics Systems Lab; Leonhardstrasse 21, 8092 Zurich, Switzerland. This research was conducted as part of the Autonomous River Cleanup project at ETH Zurich.
}
}

\begin{document}

\maketitle
\thispagestyle{empty}
\pagestyle{empty}

\begin{abstract}

We introduce a novel strategy for multi-robot sorting of waste objects using Reinforcement Learning. Our focus lies on finding optimal picking strategies that facilitate an effective coordination of a multi-robot system, subject to maximizing the waste removal potential. We realize this by formulating the sorting problem as an OpenAI gym environment and training a neural network with a deep reinforcement learning algorithm. The objective function is set up to optimize the picking rate of the robotic system. In simulation, we draw a performance comparison to an intuitive combinatorial game theory-based approach. We show that the trained policies outperform the latter and achieve up to 16\% higher picking rates. Finally, the respective algorithms are validated on a hardware setup consisting of a two-robot sorting station able to process incoming waste objects through pick-and-place operations.

\end{abstract}

\section{INTRODUCTION}

\begin{figure}[ht]
    \centering
    \includegraphics[width=0.45\textwidth]{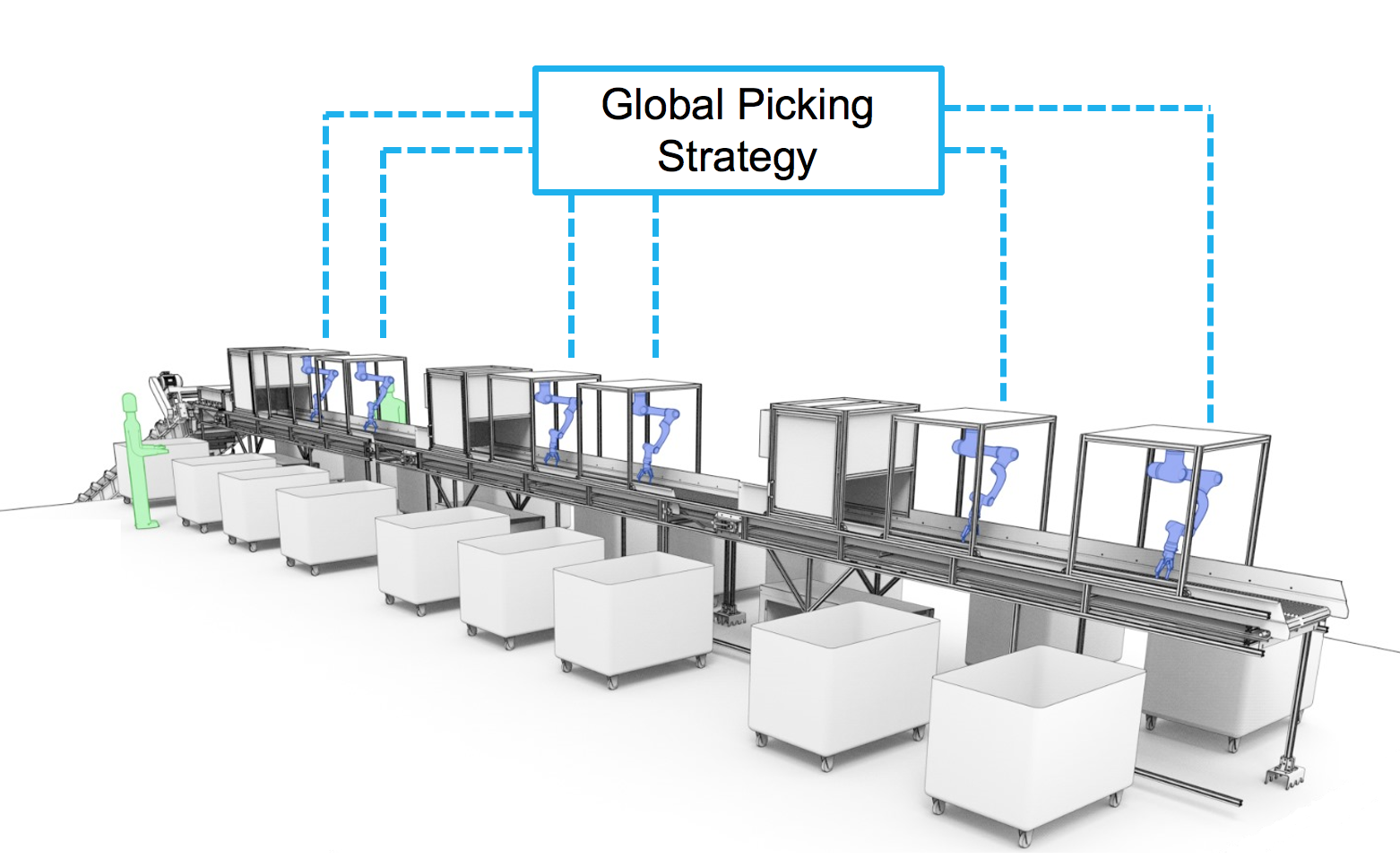}
    \caption{Artist impression of collaborative, multiple robotic arm pick-and-place setup for waste sorting, governed by a strategy to maximize picking efficiency. }
    \label{fig:advertising_image}
\end{figure}

\IEEEPARstart{P}{lastic} pollution in rivers has become a pressing global issue, with 11 million tons of plastic waste entering the ocean annually, 80\% of which is caused by 1,000 major polluting rivers \cite{c1}. To address this problem, it is desired to develop a solution capable of removing plastic and other waste objects without interfering with the existing flora and fauna essential to river ecosystems \cite{c4} . Our Autonomous River Cleanup (ARC) project, initiated in 2019, leverages robotics and automation to remove plastic waste from rivers. In order to increase the capacity at which this can be done, we enhance the existing single arm sorting station \cite{c5} with additional robot arms. For multiple robot agents to efficiently sort waste on a conveyor belt, we develop and evaluate novel strategy algorithms using reinforcement learning that assign pick-and-place (PnP) tasks to the respective robot agents (Figure \ref{fig:advertising_image}).

Given a set of objects on the moving conveyor belt, the robot agents are tasked with removing waste objects, whilst bio-matter is ignored and collected at the end of the belt. The challenge is to allocate each robot optimally with PnP operations for objects within its reachable workspace.

\section{Related Work}
Finding suitable solutions to this problem has been investigated extensively over the last decades as these technologies can be applied to several use cases such as the food industry \cite{c6}, logistics \cite{c7} and sorting recyclable packaging \cite{c8}.

As with our robotic sorting station, a single robot arm is able to fulfill the sorting task. However, a well-controlled team of robots may be preferable, as the production throughput may be considerably improved. Previous studies employed genetic algorithms to tackle multi-robot assembly planning problems, optimizing rendezvous locations for component pickup and placement \cite{c9}. Other studies have explored techniques for distributing multiple mobile robots to efficiently service various points of interest, often focusing on manufacturing tasks such as inspection, parts replacement, and data collection \cite{c10}. More recently, as the role of robots in automated manufacturing systems continues to expand, the complex task of designing and managing multi-robot material-handling systems has gained prominence, with a focus on optimizing system throughput, minimizing cycle times, and reducing work-in-progress and transitory periods \cite{c11}.

\subsection{Traditional multi-robot coordination}
\label{sec:traditional}

The traditional approach is to assess the performance of the system with a single global objective function. Studies show that the development of approaches and algorithms based on this perspective has been awarded with a substantial amount of research \cite{c12}. Prior studies employed various approaches for multi-robot task allocation, showcasing their effectiveness in specific contexts, but were limited by their failure to address proper generalized collaboration between robots \cite{c13}. In the context of multi-robot task allocation, the problem is framed as the assignment of a set of operational tasks to a group of robots, seeking to optimize the overall cost function. 

This is done while also taking into account the individual performance of each robot agent, an aspect often overlooked in previous research as only a single objective function was taken into account \cite{c14}. Other studies in dynamic task allocation relied on coordination algorithms that employed local sensing without direct communication between robots, demonstrating robustness and scalability, yet frequently lacked a comprehensive global view of the system to enhance performance \cite{c15}. Another approach presented in \cite{c16} employs dynamic programming techniques and additional heuristics. Here, the optimal object-picking sequences are computed for a predetermined finite horizon. This offers limited adaptability to most real world applications, as the task duration is often unknown.

Alternatively, in the realm of non-cooperative game theory instead of optimizing a single objective, a set of objective functions are addressed simultaneously. In this case, each agent is given the freedom of individual decision based on its considerations with respect to its assigned task \cite{c17}. In general, different results are achieved with this method compared to optimizing for a single objective function even if the global reward is the sum of each agent's contribution. This stems from the fact that the individual objective function is dependent on the action of its neighboring robots and is therefore not completely decoupled \cite{c18}.

Another approach to solving the multi-robot coordination problem is to utilize common single-player strategies such as first-in-first-out (FIFO), shortest-processing-time (SPT), and others, and finding the best combination of these part dispatching rules for a set of robot agents \cite{c19}. This is achieved with the help of a greedy search algorithm, and robustness to pattern variation is added by applying a Monte Carlo Strategy. Due to the low complexity, intuitive nature, and good performance of this method, a modified version was developed as part of this work and will serve as a suitable baseline to compare how the learning-based method performs. Lastly, the limitation of this approach becomes evident in its lack of communication among robots once the strategy combination has been selected, resulting in a lack of awareness regarding the state of their fellow robots. Recognizing the potential value of this information, our proposed reinforcement learning-based approach is designed to harness it to enhance performance.

\subsection{Learning-based multi-robot coordination}
\label{sec:learning}

Reinforcement Learning (RL) has quickly risen in popularity in a large variety of fields as a means for agents of any sort to learn new behaviors in an enclosed environment. Recent advances in deep RL have enabled the creation and use of powerful multi-agent systems, such as for the use case of multiplayer games \cite{c20} and autonomous cars \cite{c21}. Hence, there is a strong interest in extending these concepts to the use case of multi-robot coordination. A recent feasibility study has been conducted on this, which highlights the potential benefits of implementing RL algorithms \cite{c22}. The works of \cite{c23} and \cite{c24} have demonstrated the successful application of Q-learning-based algorithms to multi-agent cooperation problems. The goal here is to proficiently allocate PnP tasks. More specifically, the authors considered multi-agent domains that are fully cooperative and partially observable. This means that no agent is aware of the complete state of the environment, but rather has access to a private observation correlated to the agent's immediate surrounding and current state. Nevertheless, all agents are attempting to maximize the global objective comprised of the sum of the disjointed single rewards.

In this work, we present a reinforcement learning-based approach to coordinate multiple robot agents for a PnP tasks. This is achieved by training the model in a custom environment with proximal policy optimization \cite{c245}. The methods used for this approach together with the underlying problem formulation and assumptions will be presented in chapter \ref{methods}. Results and performance comparisons to proven methods can be found in chapter \ref{results} and are discussed in chapter \ref{discussion}.

\section{METHODS}
\label{methods}

\subsection{Problem formulation and assumptions}

\begin{figure}[ht]
    \centering
    \includegraphics[width=0.45\textwidth]{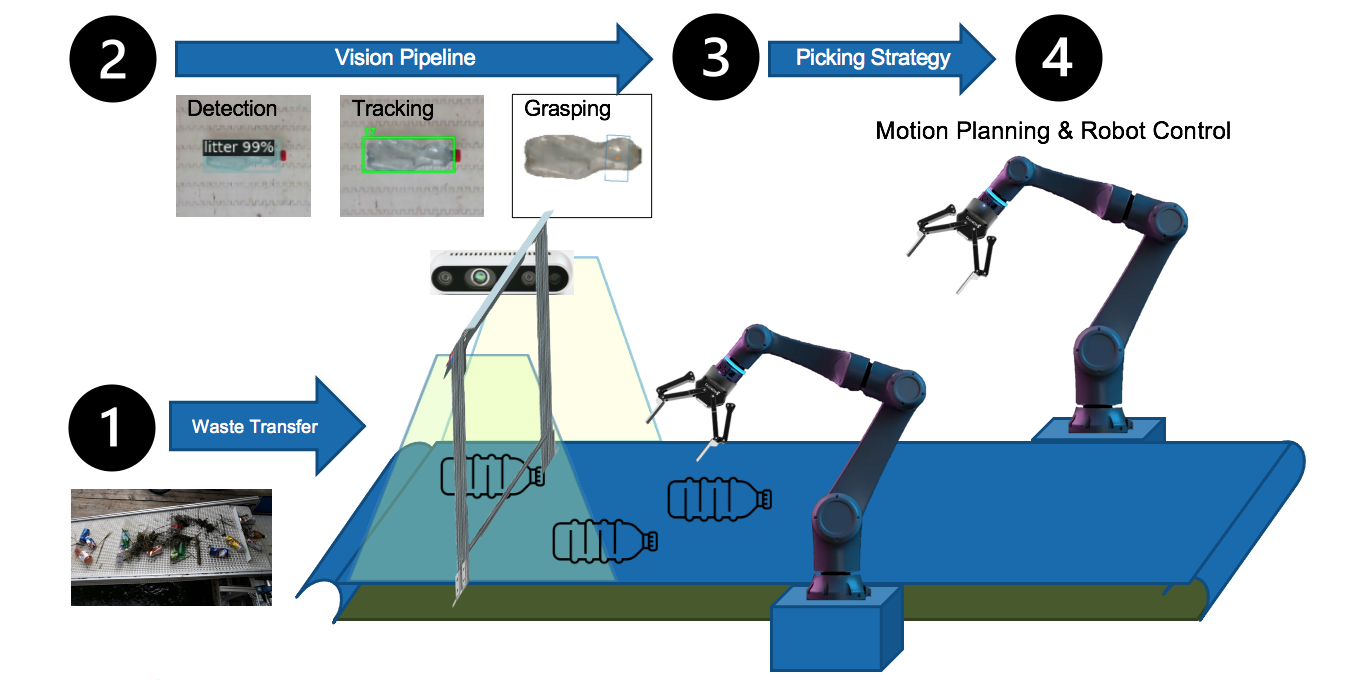}
    \caption{Pipeline on the real system consisting of waste transfer; vision pipeline to detect, track and grasp waste objects; the picking strategy assigning PnP tasks to robot agents and a motion planning and robot control module for executing the desired robot motions.}
    \label{fig:pipeline}
\end{figure}

The key requirements for optimal picking strategies can be formulated as follows. Firstly, the methods in question should be applicable to the use case of two or more robot agents. Furthermore, it is aimed to maximize the waste removal potential whilst being robust to pattern variation. A pattern refers to the arrangement of different waste objects being transported on the conveyor belt, which are to be picked up by the robots. In the simulation, this is interpreted as a moving point on the conveyor belt, in the real system this could be a waste object such as a plastic bottle. The PnP sorting algorithm should achieve high picking rates, no matter how the incoming waste object distribution looks like. Lastly, the developed solution should be directly deployable within the pipeline of the real system, which is visualized in Figure \ref{fig:pipeline}. In order to solve the multi-robot coordination problem adhering to the aforementioned requirements, a set of assumptions were made:
\begin{itemize}
    \item The multi-robot system consists of identical robot agents. They will be lined up on both sides of the conveyor belt in alternating fashion.
    \item Robot agents have non-overlapping workspaces. This means that collision avoidance between robot agents is not considered. This is a reasonable assumption since sorting waste objects is usually done in non-confined spaces. There are, however, suitable motion planner solutions available that would enable some degree of overlapping workspaces \cite{c25}.
    \item Each robot agent has a single drop-off location. It serves as the resting position for the robot agents and marks the bin location.
    \item The state of the system is fully observable. When a strategy decision has to be made, each robot can make use of the entire information available within the system in its current state.
    \item The conveyor belt moves at constant speed and objects are not sliding.
\end{itemize}

\subsection{Pattern modelling}

In order to thoroughly assess and train the methods that will be discussed in the upcoming sections, it is essential to have a multitude of different pattern distributions available. Due to the limited access to real-world data, additional random distributions need to be generated.

\begin{figure}
  \centering
  \subfloat[Poisson Disk Pattern Distribution.]{\includegraphics[height=0.16\textwidth]{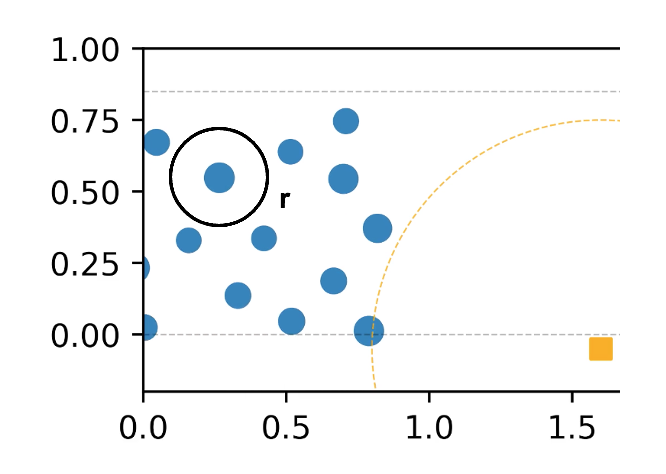}\label{fig:poisson}}
  \hfill
  \subfloat[Grid Sampled Pattern Distribution.]{\includegraphics[height=0.16\textwidth]{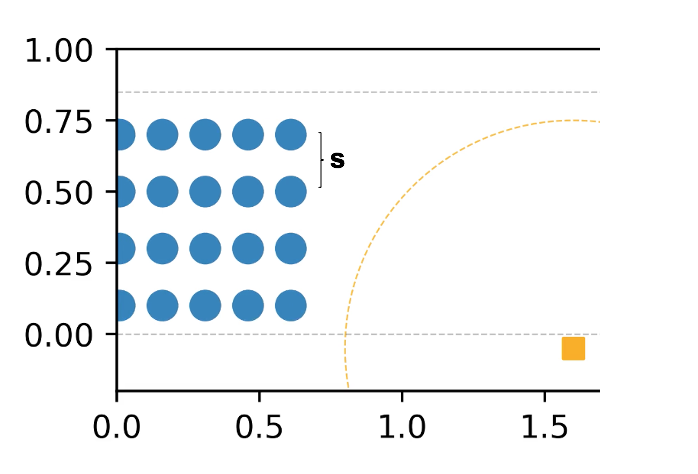}\label{fig:grid}}
  \caption{Overview of the two different pattern modelling methods. Waste objects are indicated in blue and the robot position and range are shown in orange.}
  \label{fig:distributions}
\end{figure}

The first pattern distribution type can be seen in Figure \ref{fig:poisson}. The parameter $r$ defines the minimal radius defining a circle around each object in which no other objects are spawned. This results in a spread out distribution which closely resembles what can be found on our experimental conveyor belt with real-world objects. The second type of pattern distribution can be found in Figure \ref{fig:grid}. In this case, the parameter $s$ is used to describe the spacing of the grid lattice. Even though this type of distribution is not representative for sorting waste objects, it often serves as a good setup in literature \cite{c16} to compare how well algorithms perform and was chosen for this reason.

\subsection{Combinatorial game theory (GT) strategy}

Following the implementation of \cite{c19} the concept of this approach is to interpret the robot agents as players in a non-cooperative game in which global performance is optimized. Each player is assigned a static rule, which determines how the behavior of the respective robot agent will look like over the full run time of the system. The goal is to find the best behavior rule for each robot, in order to maximize a single global objective function.

These rules can be described as single-player strategies, as they dictate the behavior of a single robot agent. They are based on different parameters such as the positioning of the waste object on the conveyor belt, the estimated processing time as well as the weight factor of a particular waste object. The most commonly used single-player strategies which were considered in this work are listed below:

\begin{itemize}
    \item First-In-First-Out (FIFO) picks up the item that has progressed the furthest on the conveyor belt.
    \item Shortest Processing Time (SPT) picks up the item with the shortest time to process.
    \item Longest Processing Time (LPT) picks up the item with the longest time to process.
    \item Shortest Distance (SD) picks up the item with the shortest distance to the robot base at the start of the PnP operation.
    \item Longest Distance (LD) picks up the item with the longest distance to the robot base at the start of the PnP operation.
    \item Priority Picking (PP) picks up the item with the highest priority based on a predefined weighting factor.
\end{itemize}

It is important to note that this list is non-exhaustive and can be further expanded. Furthermore, the concepts of these strategies can be combined with different weighting factors to generate new strategies.

Extending this concept to a multi-robot setting, the question boils down to finding an appropriate combination of strategies. This consists of assigning a single-player strategy to each individual robot. The difficulty lies in finding a suitable strategy combination whilst satisfying the key requirement of being as robust to pattern variation as possible. We use a Monte Carlo strategy for estimating the performance, while a Greedy Randomized Adaptive Search Procedure (GRASP) is used for finding the best strategy combination \cite{c29}. The latter consists of constructing an initial solution and launching a local neighborhood search afterward. The solution itself consists of a list of strategies, assigning a specific strategy behavior to each individual robot agent. After convergence in the local neighborhood search, we obtain the final strategy combination.

\subsection{Reinforcement learning strategy}

In our approach, we reformulate the multi-robot picking problem as an RL problem. Here, the action for each robot agent is defined as the next object to be collected through a PnP task. In simulation, the robot agents can take actions corresponding to picking up waste objects through PnP tasks. 

The decision of which object the robot should pick next correlates to the actions that are used to facilitate the training process. With a trained system, the resulting policy can be directly used for predicting near-optimal actions per agent given the state of the system.

More specifically, the methods of RL consider the interaction of a learning agent with its environment. The latter is formulated in discrete time as a Markov Decision Process (MDP). The environment can be described as a known state $s_t$ in the set of possible states $S$ at every time step $t$. There, the agent samples an action $a_t$ in the action space $A$ from a stochastic policy $\pi(a_t|s_t)$. Subsequently, the agent receives a reward $r_t \in R$ and moves to the next state $s_{t+1} \in S$. $R$ represents the reward space which is defined based on fulfilling certain conditions specified by the user. As we consider the system to be fully observable, the observations are available and written as $o_t = s_t$. The goal of the agent is then to find an optimal policy $\pi^*$ that maximizes the expected discounted reward over a finite time horizon. This can be formulated as follows

\begin{equation}
\label{equ:equ_1}
\pi^* = \operatorname*{argmax}_{\pi} \mathbb{E}\left[ \sum_{t=0}^{\infty} \gamma^t r_t \right]
\end{equation}

where $\gamma$ is the discount factor and was chosen to be $0.99$ as commonly done.

\subsubsection{Setup of custom environment}

In order to train an RL policy, first a custom environment of the multi-agent sorting station has to be set up. We achieve this through the custom environment capabilities of OpenAI's gym \cite{c30}. Their implementation allows users to formulate various kinds of problems as games, on which agents can be trained. Subsequently, we model an abstracted version of the sorting station as such an environment. The environment is simulated in 2D and coexists with the conveyor belt plane. A single RL policy is afterward trained to find optimal strategies.

The environment is first initialized by the size of the conveyor belt, the number of agents present, as well as with their absolute positioning on the 2D map. Furthermore, the following meta parameters can be specified: conveyor belt speed, frequency at which the system updates, robot specifications (maximum reach, maximum end-effector speed) as well as the respective bin locations. The setup is concluded by loading a pattern distribution of the user's choice and placing it such that the first object is at the start of the conveyor belt.

\subsubsection{Actions, observations, and rewards}

Thereafter, the system enters its step function. The items on the conveyor belt are progressed according to the aforementioned specifications. Once items have entered the extended workspace of the first robot, the first decision needs to be taken. At these time instances the actions, observations, and rewards need to be assigned. Note that the extended workspace refers to the area in which items are on the conveyor belt, such that after the time it took the end-effector to move to the meeting point, the items are in the real workspace. The meeting point is where the pickup operation of the gripper takes place.

The actions space assigns all objects in question for picking either a $0$ or a $1$, where the latter corresponds to picking up the item. For this reason, only one suitable object will be assigned $1$, all others $0$. In this implementation, the number of objects to be considered and hence the size of the action space is a tunable parameter. Whilst a low value results in a narrow action space with little freedom, a value chosen too large results in an overflow of choices. The policy then takes substantially longer to train and convergence is achieved at a slow rate. Through experimentation on the two-robot system with the set of parameters corresponding to the real system, the optimal value for the size of the action space was found to be $10$ items. This means the model takes 10 items into account when making decisions about which item to pick next.

Each object which is part of the actions space delivers object-specific observations and consists of the relative positioning $x_{rel}, y_{rel}$ w.r.t. their robot agent, the estimated processing time $t_{process}$, and associated object reward $r$. Furthermore, robot-specific observations are taken into account. These consist of whether an agent is currently busy with a PnP task, as well as the estimated time until the agent is available again. It is important to note here that the information used in the observation space is also available on the real system, therefore enabling a straightforward sim-to-real transition.

The first type of reward is directly gained when picking up an item, and hence object-specific. Assigning different rewards to objects opposed to uniform values serves as a tool for the user to specify how the various waste objects should be prioritized. In this implementation, multiple considerations were taken into account. Firstly, the volume and mass metric of a waste object is in direct correlation with the amount of recyclable material that is retrieved. 

For the implementation on the real system, this is estimated based on the 2D area $A$ which is covered by the waste object during the vision pipeline. Furthermore, the certainties for the waste detection score $p_{detection}$ as well as the grasp prediction score $p_{grasp}$ are further used for the final object-specific reward. The reasoning behind this is that a higher probability of retrieving real waste objects as well as achieving a successful grasp should be prioritized. 

Lastly, the range of the robot-specific rewards was set to be between $0$ and $1$, hence, a sigmoid activation function was implemented. The final equation used in this work to compute the object specific reward $r$ is formulated as

\begin{equation}
\label{equ:equ_2}
r = p_{detection}*p_{grasp}*sig(k*A)
\end{equation}

where $k$ is a tuning parameter and was tuned to $1/100$.

Secondly, a fixed reward is achieved once a \textit{done} state of the system has been reached. This is the case if either all items were picked up or once the last object has left the conveyor belt. Once thereafter, another reward is assigned in correspondence to the object-specific reward-weighted picking rate.

\begin{table}[tb!]\centering
\resizebox{\linewidth}{!}{
\begin{tabular}{rlrl} \\ \toprule
    Environment & & PPO &\\
    \midrule
        Size of action space & 10 & Total Epochs &  3M\\
        Simulation frequency & 10 Hz & Steps per update & 2048\\
        Conveyor belt speed & 0.078 m/s & Batch size & 64\\
        Max end effector speed & 0.5 m/s & Discount factor & 0.9995\\
        Max robot range & 0.8 m & Learning rate & 0.0003\\
 \bottomrule
\end{tabular}
}
\caption{Hyper-parameters used for training. All other PPO parameters are kept at the default values of the stable baselines' implementation.}
\label{table_hyper_param}
\end{table}

\subsubsection{Training with proximal policy optimization}

Policy gradient methods are fundamental in the recent breakthroughs of applying RL to solve problems in the field of control theory, video games, or board games such as Go \cite{c31}. 

This, together with the recent success of RL in the field of robotics, motivated the choice to use Proximal Policy Optimization (PPO) for training. \\

The key benefit of PPO is that it was designed to prevent destructively large policy updates during training. It is able to do so in a computationally efficient manner, and therefore has become one of the default choices for RL algorithms. The implementation of the algorithm was provided by the open-source StableBaselines3 library \cite{c32}. For training, the custom sorting station environment is vectorized to increase the speed at which the learning is achieved. 

Furthermore, it involves a learning curriculum starting with sparser pattern distributions to which optimal strategies can be found more rapidly and slowly increasing in difficulty by considering denser pattern distributions. The set of tuned hyper-parameters can be found in Table \ref{table_hyper_param}.

\section{RESULTS}
\label{results}

\subsection{GT approach}
\label{section:gt_results}

The combinatorial GT approach was applied to different sorting station setups with a varying number of robot agents. The setups were tested on a balanced set of pattern distributions following Poisson and grid sampling. The 120 Poisson distributions were sampled with the key parameter $r$ set to be between $0.15$ and $0.4$. The grid sampled patterns were achieved with a lattice spacing $s$ of the same range. Furthermore, the conveyor belt speed was set to $0.05$ m/s. The resulting strategy combination solutions can be seen in Table \ref{table_1}.

\begin{table}[h]
\begin{center}
 \begin{tabular}{||l c c c||} 
 \hline
 \! & Patterns & Distribution Types & Final Solution \\ [0.5ex]
 \hline\hline
 2 Agents & 240 & Poisson+Grid & [SPT; FIFO] \\ 
 \hline
 3 Agents & 240 & Poisson+Grid & [SPT; SPT; FIFO] \\
 \hline
 4 Agents & 240 & Poisson+Grid & [SPT; SPT; SPT; FIFO] \\ 
 \hline
 \end{tabular}
 \caption{\label{table_1} Most robust strategy combinations in terms of the tested pattern distributions maximizing the pick rates of the overall system resulting from the combinatorial game theory approach.}
\end{center}
\end{table}

As depicted above, the robust strategy combination for a two robot agent sorting setup is [SPT; FIFO]. This means that the first agent is assigned the SPT rule and the second agent follows FIFO. This result can be intuitively explained as follows. The first agent acts as the "hard worker" and prioritizes objects which take the least amount of processing time. As such, it achieves a high picking rate whilst certain objects are neglected and move outside the reachable workspace. To compensate for this, the second robot prioritizes objects that have progressed the furthest along the conveyor belt. Together, these two single-player strategies form an efficient strategy combination for a two-agent setup. This strategy combination is found as the optimal solution when a close to 100\% picking rate can be achieved. However, if the conveyor belt is too crowded with objects, the algorithm will propose to assign both robots to the SPT rule.

A similar behavior carries over when increasing the number of agents that are involved during sorting. For three and four agents respectively, every robot agent except for the last one follows SPT, whilst the last one functions as the leftover picker through FIFO. This can be similarly explained as previously, the only thing that changes is the number of ``hard workers''. It is important to note here that the work of \cite{c19} came to a similar conclusion. Whilst these solution combinations are dependent on the sampled pattern distributions and therefore inherently biased, they still serve as a good baseline to compare the RL approach with.

\subsection{Comparison of GT and RL approach}

The RL approach has been implemented to accommodate as many robot agents as the user wishes to specify. Due to the fact that the real system consists of two robot agents, the focus was set on such a configuration. The model was trained for a total of 3 million epochs which takes around two to three hours of computation time to achieve the final policy\footnote{All computations for the results were run on a 2015 MacBook Pro, 2,7 GHz Dual-Core Intel Core i5 Processor running Ubuntu 18.04.}. The following results compare the robust combinatorial solution (optimized for a set of pattern distributions) versus the RL approach, as well as the greedy combinatorial approach (optimized for the single pattern distribution in question). The latter is achieved by finding the best strategy combination solely based on the single pattern on which it is tested on. It therefore always achieves equal or better performance compared to the first one, whilst losing the robust nature.

\begin{table}[h]
\centering
\begin{tabular}{|| l | p{0.6cm} p{0.6cm} | p{0.6cm} p{0.6cm} | p{0.6cm} p{0.6cm} ||}
\cline{2-7}
\multicolumn{1}{c|}{} & \multicolumn{2}{c|}{Robust GT} & \multicolumn{2}{c|}{RL} & \multicolumn{2}{c||}{Greedy GT} \\
\hline
Distribution type & Picked & Time & Picked & Time & Picked & Time \\ [0.5ex]
\hline\hline
\shortstack{Grid s=0.15} & 92\% & 96.3s & \textbf{96\%} & \textbf{92.8s} & 92\% & 95.6s \\
\hline
\shortstack{Grid s=0.3} & 100\% & 113.5s & 100\% & \textbf{102.3s} & 100\% & 107.6s \\
\hline
\shortstack{Poisson s=0.2} & 96\% & 81.5s & 100\% & \textbf{70.3s} & 100\% & 74.4s \\
\hline
\shortstack{Poisson s=0.3} & 100\% & 114.8s & 100\% & \textbf{110.5s} & 100\% & 114.8s \\
\hline
\end{tabular}
\caption{\label{table_2} Comparison between robust GT (first column), RL (second column), and greedy GT (third column) solutions. Note that the pattern distributions of these four use cases were not used for training the RL policy.}
\end{table}

Table \ref{table_2} shows four different distribution patterns and the achieved results. To compare the different methods, two grids and two Poisson distributions were chosen. The corresponding parameters grid spacing $s$ and radius $r$, the percentage of objects picked as well as the time it took to do so are depicted. These test cases were designed to be realistic for the two robot sorting station.

\begin{table}[h]
\begin{center}
 \begin{tabular}{|| l | c | c | c | c ||} 
 \hline
 Distribution type & Robust GT & RL & Greedy GT & Benefit of RL \\ [0.5ex]
 \hline\hline
 Grid s=0.15 & 28.0 & 31.1 & 28.8 & \textbf{11.1\%} \\
 \hline
 Grid s=0.3 & 26.4 & 27.8 & 29.4 & \textbf{5.3\%} \\
 \hline
 Poisson s=0.2 & 21.3 & 24.8 & 23.3 & \textbf{16.4\%} \\
 \hline
 Poisson s=0.3 & 18.7 & 19.1 & 18.7 & \textbf{2.1\%} \\
 \hline
 \end{tabular}
 \caption{\label{table_3} Achieved picking rates in [picks/minute] for the four use cases, as well as the benefit of using RL over robust GT.}
\end{center}
\end{table}

The RL approach outperforms the other two presented options in most use case. Even if the GT approach manages to pick up all items, the RL approach manages to achieve the same result within a shorter amount of time, while preserving the advantage of generalizing to various distributions. This results in an up to 16\% higher picking rates of the robots achieved through RL over the robust GT methods, as depicted in Table \ref{table_3}. The benefit of using the RL approach diminishes as we consider sparser distributions, and is also highlighted by the low picking rate value with a Poisson $s$=0.3 distribution. Here the challenge of finding optimal picking strategies decreases due to the fact that there are simply fewer choices available which could lead to more efficient results. The fewer available choices result in a lower total picking rate for any strategy. Therefore, also requiring less optimization. In conclusion, the benefit of using RL is most prominent when there are a lot of waste objects present on the conveyor belt system.

\subsubsection{Final multi-robot sorting station}

\begin{figure}[h]
   \centering
   \includegraphics[width=0.45\textwidth]{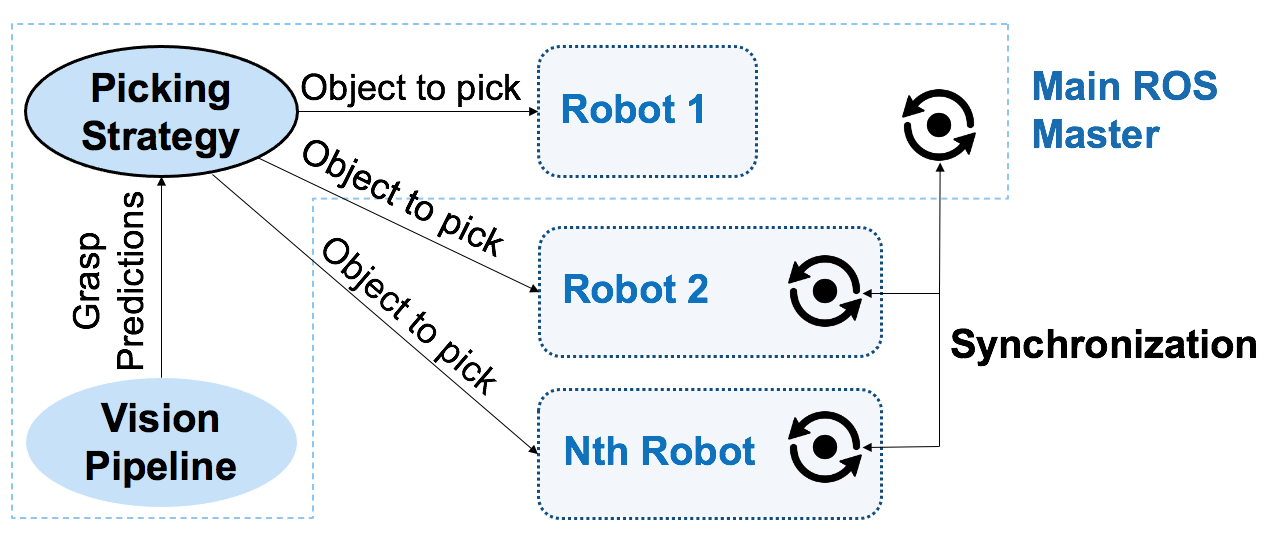}
   \caption{Overview of the general ROS architecture of the network setup.}
   \label{fig:system_architecture}
\end{figure}

\begin{figure}[ht]
    \centering
    \includegraphics[width=0.45\textwidth]{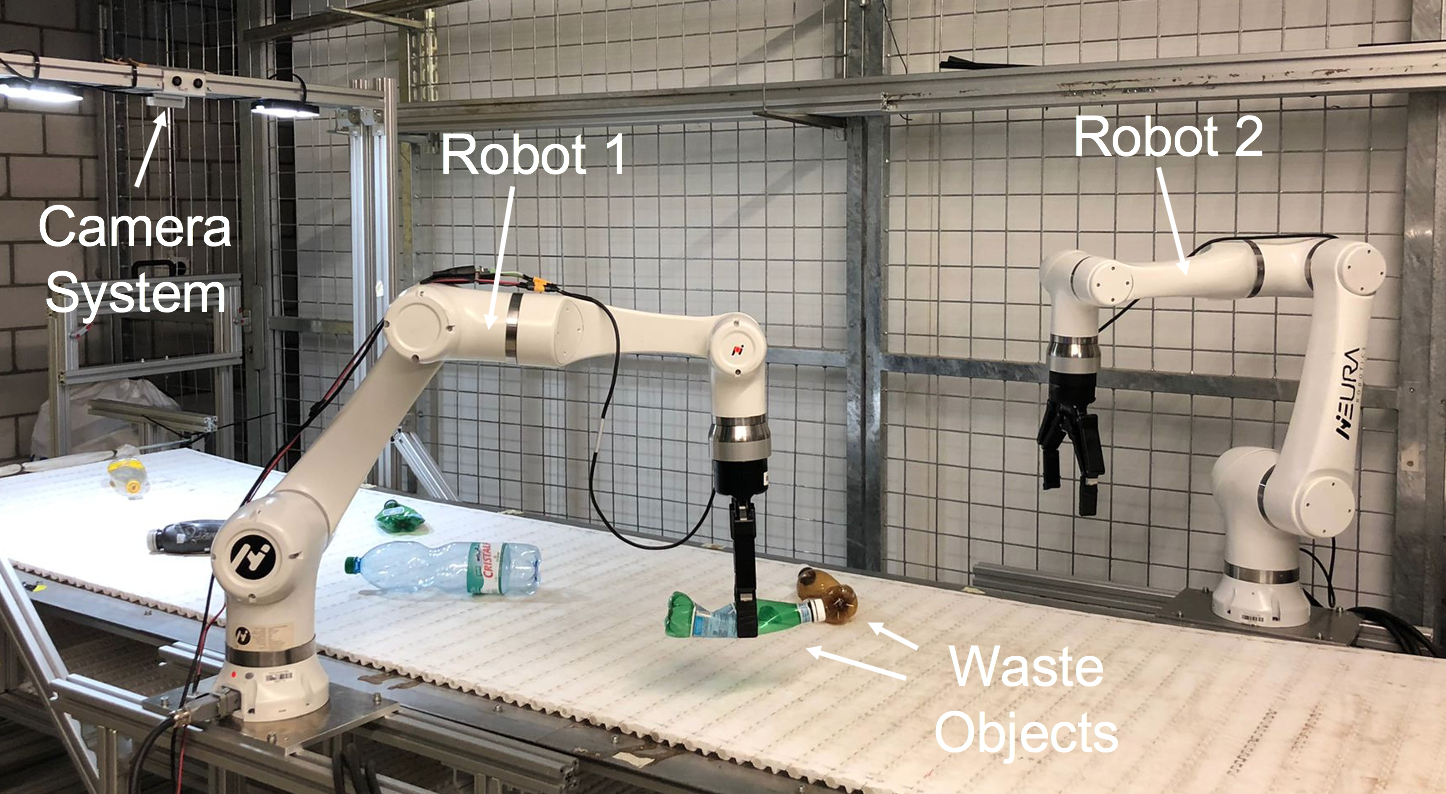}
    \caption{Final, two-robot sorting station handling incoming waste objects. One cycle of a single robot agent is achieved as follows: the robot is waiting in its resting position for the next object to pick. After the picking strategy decision has been made, the arm is on its way to the meeting point, where the grasping maneuver is executed. After successful pickup the object is transported back to the starting position where it is dropped off.}
    \label{fig:final_system}
    \vspace{-1em}

\end{figure}

Both the combinatorial GT and RL approach have been deployed on the real system to showcase that a transfer from simulation is feasible. We introduce the picking strategy following the block diagram seen in Figure \ref{fig:system_architecture}. The picking strategy node receives the required information about the environment from the vision pipeline and sends high level PnP tasks to the respective robot nodes. The two robot sorting station is fully operational and sorting waste objects with multiple robot agents is enabled as can be seen in Figure \ref{fig:final_system}.

\section{DISCUSSION}
\label{discussion}
\subsubsection{Influence on conveyor belt speed}
    
Whilst the above test results focus on the achieved picking rates, it is also of interest to consider the benefit of the RL approach on the conveyor belt speed. More specifically, the question arises as to what the maximum conveyor belt speed is, such that all waste objects can still be picked. For this, the grid distribution with $s$=0.15 is considered, as it simulates a typical and highly demanding situation for the ARC sorting station. The maximum speed of the conveyor belt is then found to be \SI{0.042}{\meter\per\second} and \SI{0.047}{\meter\per\second} for the robust GT and RL approach, respectively. Therefore, during high demanding situations, the system can be run with an 11.9\% faster conveyor belt speed when utilizing the presented RL method, resulting in more throughput.

\subsubsection{Advantages of the RL approach}

The proposed RL approach differentiates itself from previous efforts in solving similar task allocation algorithms by its flexible nature. It will not get stuck in one fixed strategy, but rather has the ability to flexibly assess each situation for a currently \textit{best} strategy. In addition, the proposed method can capture various kinds of complexities by tuning observations, actions and rewards. The observation space allows for direct interaction with the state of the system, thus the possibility to add more information if desired. Similarly, to increase complexity in behavior we can increase the action space by adding options such as for example flexible end-effector resting positions, throwing behavior or bin areas instead of bin points. With increasing observations and actions available, it is then of importance to tune rewards accordingly such that the desired behavior is achieved with the trained policy.

\subsubsection{Scalability}

Future work in regard to the presented work includes investigating how well the RL approach scales with more than two agents. It is reasonable to assume that better results can be achieved in that scenario as well. Another point that could be of interest is splitting up the global picking strategy policy into robot specific policies. This could further help running systems with more than two agents without running into computational scalability issues. The question then stands which information exchange between robot agents is necessary such that the main benefit of global awareness and collaboration is maintained.

\section{CONCLUSION}

We demonstrated a novel method for multi-robot sorting using Reinforcement Learning. For this, we formulated the picking strategy problem in simulation and used Poisson disk and grid sampled modeling concepts for waste pattern distributions. Novel reinforcement learning-based as well as proven combinatorial game theory-based strategies were developed and implemented in the simulation. For the two robot setup, a trained RL policy consistently outperformed the combinatorial approach on varying object distributions, achieving up to 16\% higher picking rates. The advantages of the RL method shows especially good results in dense distributions of objects to be picked. Proof of concept tests have been successfully executed on a two-robot hardware setup, showing the applicability to real-world scenarios.

\section*{ACKNOWLEDGMENT}

The authors would like to thank the various foundations and partners that support the efforts of the Autonomous River Cleanup project, most notably the D{\"a}twyler Stiftung, R{\"u}tli Stiftung, Fondation Valery, ewz naturemade star Fonds, LIFE Klimastiftung, Somaha Stiftung and Neura Robotics (see riverclean.ethz.ch/ for a detailed list of supporters).

\end{document}